\documentclass[letterpaper]{article} 
\usepackage{aaai2026}  
\usepackage{times}  
\usepackage{helvet}  
\usepackage{courier}  
\usepackage[hyphens]{url}  
\usepackage{graphicx} 
\usepackage{natbib}  
\usepackage{caption} 
\usepackage{adjustbox}
\usepackage{algorithm}
\usepackage{algorithmic}
\usepackage{threeparttable}
\usepackage{multirow}
\usepackage{booktabs}
\usepackage{xcolor}
\usepackage{newfloat}
\usepackage{listings}
\usepackage{amsmath}
\usepackage{amssymb} 
\DeclareCaptionStyle{ruled}{labelfont=normalfont,labelsep=colon,strut=off} 
\floatstyle{ruled}
\newfloat{listing}{tb}{lst}{}
\floatname{listing}{Listing}
\title{PMPGuard: Catching Pseudo-Matched Pairs in Remote Sensing Image–Text Retrieval}
\author{
    Pengxiang Ouyang\textsuperscript{\rm 1},
    Qing Ma\textsuperscript{\rm 2}\thanks{Corresponding Author}, 
    Zheng Wang\textsuperscript{\rm 1},
    Cong Bai\textsuperscript{\rm 1,3},
}
\affiliations{
    \textsuperscript{\rm 1}College of Computer Science and Technology, Zhejiang University of Technology, Hangzhou 310023, China \\
    \textsuperscript{\rm 2}School of mathematical sciences, Zhejiang University of Technology, Hangzhou 310023, China \\ 
    \textsuperscript{\rm 3}Zhejiang Key Laboratory of Visual Information Intelligent Processing, Hangzhou 310023\\ 
    oypx@zjut.edu.cn, maqing@zjut.edu.cn, zhengwang@zjut.edu.cn, congbai@zjut.edu.cn 
}

\begin{document}

\maketitle

\begin{abstract}
Remote sensing (RS) image–text retrieval faces significant challenges in real-world datasets due to the presence of Pseudo-Matched Pairs (PMPs), semantically mismatched or weakly aligned image–text pairs, which hinder the learning of reliable cross-modal alignments. To address this issue, we propose a novel retrieval framework that leverages Cross-Modal Gated Attention and a Positive–Negative Awareness Attention mechanism to mitigate the impact of such noisy associations. The gated module dynamically regulates cross-modal information flow, while the awareness mechanism explicitly distinguishes informative (positive) cues from misleading (negative) ones during alignment learning. Extensive experiments on three benchmark RS datasets, i.e., RSICD, RSITMD, and RS5M, demonstrate that our method consistently achieves state-of-the-art performance, highlighting its robustness and effectiveness in handling real-world mismatches and PMPs in RS image–text retrieval tasks.
\end{abstract}

\section{Introduction}
Remote-sensing (RS) image--text retrieval, which aims to establish accurate semantic correspondence between aerial imagery and natural-language descriptions~\cite{DBLP:journals/inffus/LiMZ21, DBLP:conf/cits/QuLTL16}, underpins a broad range of Earth-observation applications---from disaster monitoring and land-use classification to urban planning. The task is inherently cross-modal: given a textual query, one must retrieve the most relevant RS images, and vice versa. Yet several challenges impede progress. First, the visual and textual domains exhibit a pronounced modality gap, exacerbated by the rich, fine-grained semantics typical of overhead scenes~\cite{DBLP:conf/icmcs/OuyangCMWB24, DBLP:conf/cvpr/LinWQMCB25, DBLP:journals/tgrs/OuyangMB25}. Second, existing RS datasets are often constructed via automated captioning or legacy metadata pipelines, resulting in incomplete, error-prone, or semantically extraneous descriptions. Consequently, a substantial fraction of image-text pairs are only partially aligned-or outright mismatched-introducing noisy supervision that misleads models trained under the assumption of perfect correspondence. To sustain robust retrieval in the face of such imperfect annotations, it is essential not only to suppress the deleterious effects of pseudo-matched pairs while distilling potentially informative cues from them~\cite{zhang2024rs5m}.
\begin{figure}[t]
	\centering
	\includegraphics[width=1\linewidth]{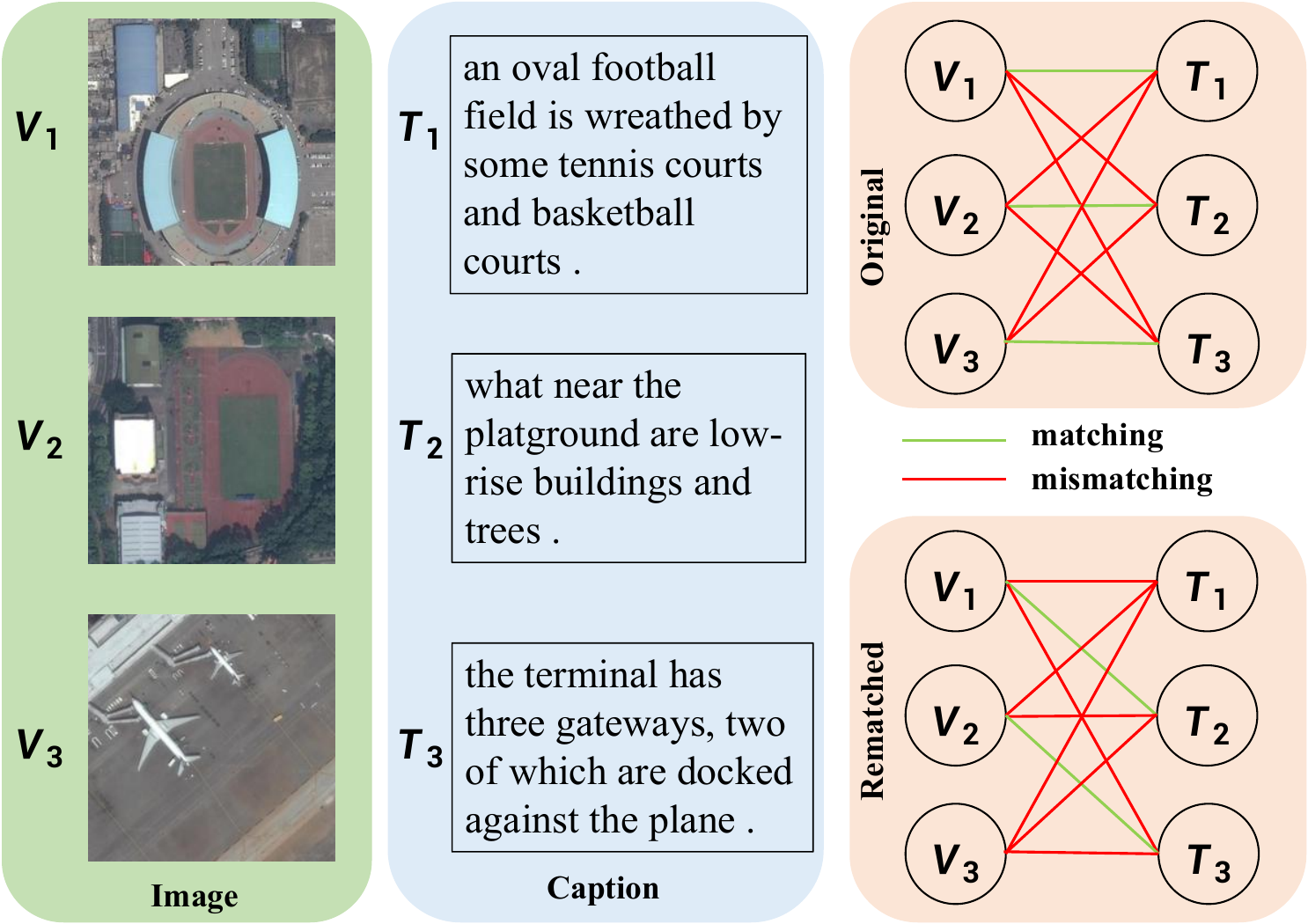}
	\caption{A simple example illustrates our key insight: Pseudo-Matched Pairs (PMPs), image–text samples with partial or incorrect alignment, are not merely noise but often contain latent semantic cues. Instead of discarding them, PMPGuard exploits these cues by rematching semantically relevant pairs (green links) and repelling irrelevant ones (red links), effectively transforming noisy supervision into useful alignment signals.}
	\label{fig1}	
\end{figure}

\begin{figure*}[t]
	\centering
	\includegraphics[width=1\linewidth]{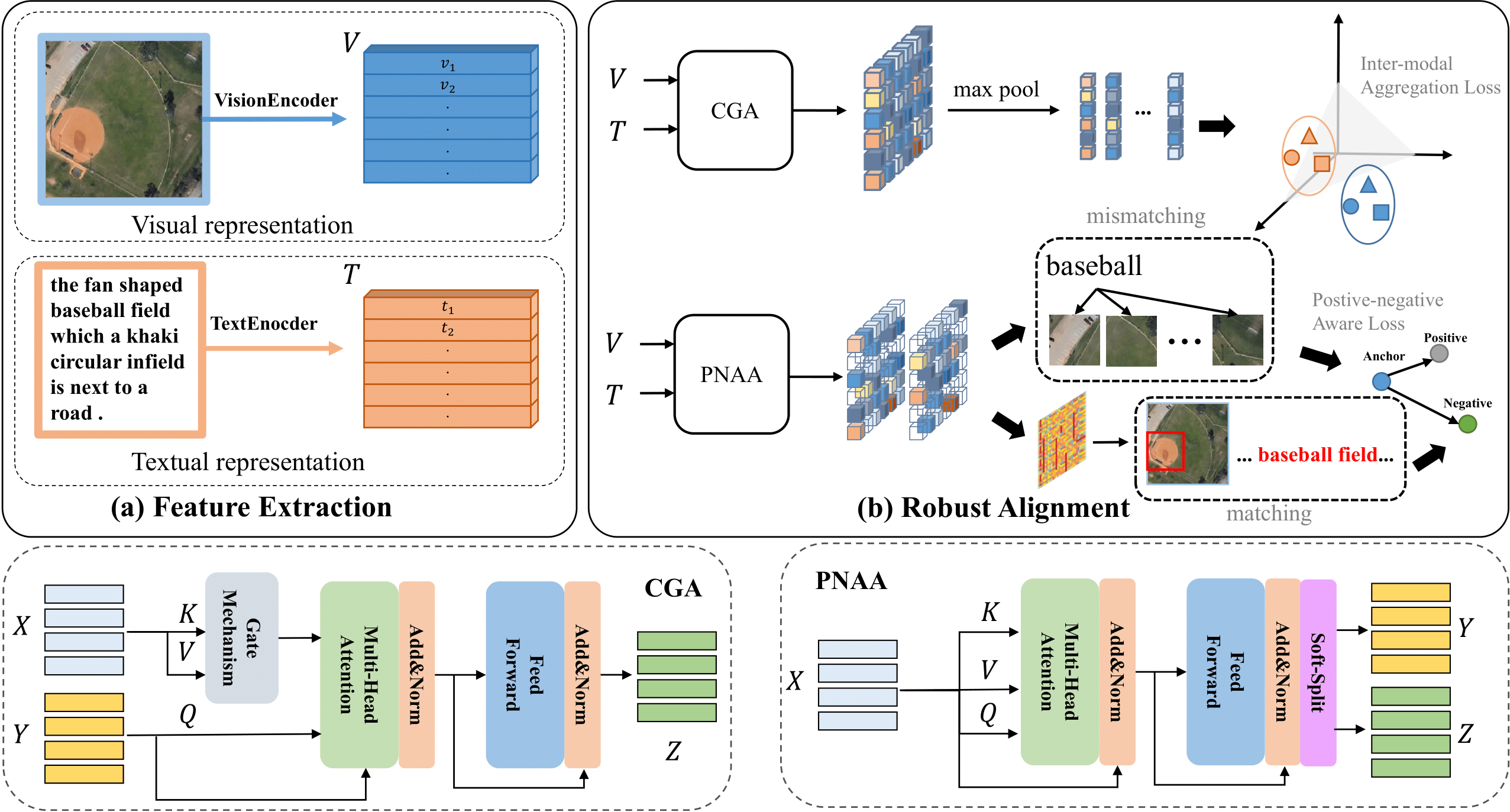}
	\caption{PMPGuard overview: vision and text encoders extract features, then Cross-Gated Attention (CGA) suppresses mismatched cues and Positive–Negative Awareness Attention (PNAA) contrasts reliable/unreliable pairs, jointly optimized to mitigate pseudo-matched pairs and strengthen cross-modal alignment.}
	\label{fig2}	
\end{figure*}
Most existing methods~\cite{DBLP:journals/tgrs/YuJ22,DBLP:journals/tip/WangPXT22,DBLP:journals/tgrs/ChenHPZCTL22} rely on well-annotated image–text pairs to learn effective joint representations. However, in the field of remote sensing, collecting high-quality and precisely aligned image–caption pairs is extremely costly and often infeasible. As a result, many datasets are constructed using web-crawled or automatically generated descriptions, which inevitably introduce Pseudo-Matched Pairs (PMPs)—pairs where the image and caption are only partially aligned or semantically inconsistent. To mitigate this issue, recent studies have attempted to down-weight the influence of PMPs or relax the loss margins in ranking-based objectives. Howeverese, these approaches tend to underexploit the mismatched samples, treating them merely as noise to suppress rather than as potentially informative signals to mine.

As shown in Fig. 1, a simple example illustrates our core idea: the potential semantic similarity between unpaired samples enables the extraction of valuable knowledge from mismatched pairs. In remote sensing image-text retrieval, such PMPs are common, where image-text samples exhibit partial semantic relevance but are not perfectly aligned. This introduces two major challenges: (1) distinguishing truly mismatched elements from partially relevant ones, and (2) leveraging useful information from mismatched pairs without being misled by noisy signals. To address these issues, we propose \textbf{PMPGuard}, a novel framework that enhances cross-modal alignment by explicitly identifying and utilizing latent semantic cues within PMPs. Rather than discarding mismatched pairs as noise, PMPGuard rematches semantically relevant cross-modal samples (green links) and repels irrelevant ones (red links), leading to more robust retrieval performance. Our main contributions are summarized as follows:

    
    
    

\begin{itemize}
    \item We propose a novel retrieval framework, \textbf{PMPGuard}, specifically designed to tackle semantic misalignment and noisy supervision caused by \textbf{Pseudo-Matched Pairs (PMPs)}, image–text pairs that are only partially or incorrectly aligned, which are prevalent in large-scale remote sensing datasets. Rather than simply suppressing PMPs, \textbf{PMPGuard} identifies and leverages them to strengthen cross-modal alignment.
    
    \item We introduce a \textbf{Cross-Gated Attention (CGA)} module that adaptively regulates cross-modal feature exchange. CGA selectively promotes semantically consistent content flow while filtering out modality-specific noise and PMP-induced mismatches during training.

    \item We design a \textbf{Positive–Negative Awareness Attention (PNAA)} module that explicitly distinguishes between aligned and misaligned regions in image–text pairs. By modeling both positive and negative cues via a dual-branch structure, PNAA improves robustness against noisy supervision from PMPs.
    
    \item Extensive experiments on three public remote sensing benchmarks, i.e., \textbf{RSICD}, \textbf{RSITMD}, and \textbf{RS5M}, demonstrate that \textbf{PMPGuard} consistently achieves \textbf{state-of-the-art (SOTA)} performance. Notably, its advantage is more evident under higher mismatch rates, confirming its robustness and generalization ability in the presence of PMPs.
\end{itemize}

\section{Related Work}
\subsection{Image-Text Retrieval in Remote Sensing}
Image-text retrieval in remote sensing (RS) seeks to align aerial imagery with natural language, enabling applications such as land cover analysis and disaster response. Recent years have witnessed a shift from traditional CNN-RNN frameworks to transformer-based models with more fine-grained alignment capabilities. For instance, PIR~\cite{DBLP:conf/mm/PanMB23} incorporates prior knowledge into the retrieval process, while SWAN~\cite{pan2023reducing} introduces scene-aware aggregation to reduce semantic confusion. DOVE~\cite{ma2024direction} enhances directional alignment using visual-semantic embedding. These methods typically assume well-aligned training pairs, which limits their robustness in noisy real-world datasets. To address imperfect supervision, recent works like BiCro~\cite{DBLP:conf/cvpr/YangXWYYL023} and DEL~\cite{DBLP:journals/tmm/FengZGLH24} propose noise-aware training strategies. However, they mostly suppress mismatches instead of leveraging them. Our approach differs by actively mining mismatched pairs through gated attention and positive-negative awareness, enabling more robust alignment under noisy correspondence.

\subsection{Mismatch-Robust Retrieval Models}
Mismatch-robust retrieval focuses on learning reliable cross-modal alignments when training data contains noisy or pseudo-matched  image-text pairs. This is especially relevant for remote sensing datasets, where large-scale annotation is often noisy or weakly supervised. DEL~\cite{DBLP:journals/tmm/FengZGLH24} introduces evidential learning to estimate uncertainty and reduce the influence of incorrect pairs. BiCro~\cite{DBLP:conf/cvpr/YangXWYYL023} enforces bi-directional similarity consistency to rectify noisy correspondences. L2RM~\cite{DBLP:conf/cvpr/HanZDL024} explicitly rematches mismatched pairs during training using a joint re-alignment strategy. Although effective, most existing methods treat mismatched pairs as noise to be suppressed. In contrast, our approach not only identifies mismatches but also exploits their latent semantic relations through gated attention and discriminative mining, enabling more robust and informative alignment~\cite{cheng2021deep}.
\section{Approach}
\subsection{Problem Definition}
Without loss of generality, we take visual--text retrieval as an illustrative task to formalize the PMP (Pseudo-Matched Pair) problem in cross-modal retrieval. Let the training set be 
\[
\mathcal{D}=\{(V_i,T_i,m_i)\}_{i=1}^{N},
\]
where $(V_i,T_i)$ denotes a visual--text pair and $m_i\in\{0,1\}$ indicates whether the pair is semantically matched.

The crux of cross-modal retrieval lies in measuring similarity across heterogeneous modalities. Existing approaches first project visual and textual inputs into a shared embedding space via modality-specific encoders $f_v$ and $f_t$, respectively, and then compute the similarity of a pair $(V_i,T_j)$ as
\[
S_{ij}=g\!\bigl(f_v(V_i),\,f_t(T_j)\bigr),
\]
where $g$ is either a parametric or a non-parametric mapping.

Most prior methods tacitly assume that every pair labeled with $m_i=1$ is indeed matched. In practice, however, real-world remote-sensing datasets—compiled through automated captioning pipelines, crowd-sourcing, or legacy metadata—frequently exhibit a non-negligible proportion of mismatched visual–text pairs that are erroneously annotated as positive. These pseudo-matched pairs (PMPs) propagate noisy supervision signals and compromise model robustness. Our objective is therefore to detect and suppress the influence of PMPs, thereby enabling robust cross-modal retrieval for remote-sensing data.

\subsection{Cross-Gated Attention Mechanism}

Addressing the noisy feature interaction problem caused by PMPs (as noted in Introduction), our Cross-Gated Attention Mechanism dynamically regulates information flow. Specifically, it solves two subproblems: (1) suppressing irrelevant features that may arise from mismatched pairs, while (2) preserving potentially useful cross-modal interactions.

Given a textual feature sequence $\mathcal{U} = \{u_i\}_{i=1}^m$ and a set of image region features $\mathcal{V} = \{v_j\}_{j=1}^n$, the goal is to compute modality-aware representations through mutual attention and gating operations.

\paragraph{Cross Attention.}
We first compute the attention score between each word $u_i$ and image region $v_j$ as:

\begin{equation}
a_{ij} = \frac{u_i^\top W_a v_j}{\|u_i\| \|v_j\|},
\end{equation}

where $W_a$ is a learnable projection matrix. The attention weights are then normalized using softmax:
\begin{equation}
\alpha_{ij} = \frac{\exp(a_{ij})}{\sum_{j'=1}^n \exp(a_{ij'})}, \\
\beta_{ji}  = \frac{\exp(a_{ij})}{\sum_{i'=1}^m \exp(a_{i'j})}.
\end{equation}
Using these attention weights, we generate cross-modal attended features:

\begin{equation}
\tilde{v}_i = \sum_{j=1}^n \alpha_{ij} v_j, \\
\tilde{u}_j = \sum_{i=1}^m \beta_{ji} u_i.
\end{equation}

\paragraph{Gating Mechanism.}
To suppress irrelevant or noisy features and highlight important interactions, we introduce gating functions that control how much information is retained from each modality. The gated representations are computed as:

\begin{equation}
g^u_i = \sigma(W^u_g [\mathbf{u}_i; \tilde{\mathbf{v}}_i] + b^u_g), \\
g^v_j = \sigma(W^v_g [\mathbf{v}_j; \tilde{\mathbf{u}}_j] + b^v_g),
\end{equation}

where $[\cdot;\cdot]$ denotes concatenation, $\sigma$ is the sigmoid activation, and $W^u_g, W^v_g$ and $b^u_g, b^v_g$ are learnable parameters. $\mathbf{u}_i$ and $\mathbf{v}_j$ are the original modality-specific embeddings, $\tilde{\mathbf{v}}_i$ and $\tilde{\mathbf{u}}_j$ are the cross-modal contexts, and $g_i^u, g_j^v \in [0,1]^d$ are the learned gating vectors.

The final cross-gated features are:

\begin{equation}
\hat{\mathbf{u}}_i = g^u_i \odot \mathbf{u}_i + (1 - g^u_i) \odot \tilde{\mathbf{v}}_i, \\
\hat{\mathbf{v}}_j = g^v_j \odot \mathbf{v}_j + (1 - g^v_j) \odot \tilde{\mathbf{u}}_j,
\end{equation}

where $\odot$ denotes element-wise multiplication. This mechanism allows the model to selectively combine intra- and inter-modal features, yielding enhanced representations for image-text matching.

\subsubsection{Inter-modal Aggregation Loss}
To align the vision and language representations after the Cross-Gated Attention (CGA) refinement, we introduce the Inter-modal Aggregation Loss (IA). Let $\hat{\mathbf{u}}_i$ and $\hat{\mathbf{v}}_j$ denote the gated visual and textual features. The loss is formulated as an InfoNCE objective with a temperature parameter $\tau$:


\begin{equation}
\mathcal{L}_{\text{IA}} = -\sum_{(i,j)\in \mathcal{P}} \log \frac{\exp\!\bigl(\frac{\hat{\mathbf{u}}_i^{\!\top}\hat{\mathbf{v}}_j}{\tau}\bigr)}
{\displaystyle\sum_{k}\exp\!\bigl(\frac{\hat{\mathbf{u}}_i^{\!\top}\hat{\mathbf{v}}_k}{\tau}\bigr)
+ \sum_{k}\exp\!\bigl(\frac{\hat{\mathbf{u}}_k^{\!\top}\hat{\mathbf{v}}_j}{\tau}\bigr)},
\end{equation}

where $\mathcal{P}$ denotes the set of semantically matched pairs. By maximizing the agreement between paired samples while pushing non-pairs apart, this loss strengthens robust cross-modal alignment and alleviates the adverse impact of pseudo-matched pairs.

\subsection{Positive-Negative Awareness Attention}
To tackle the PMPs' dual nature (both misleading and potentially useful), PNAA explicitly models positive and negative signals through: (1) a negative branch that identifies and suppresses truly mismatched fragments (Eq.12-13), and (2) a positive branch that extracts useful semantic cues from partially matched regions (Eq.14-15). This dual strategy directly addresses the challenge of utilizing imperfect training pairs raised in our problem statement.

\paragraph{Catching Pseudo-Matched Pairs}

To distinguish matched and mismatched image–text pairs, GeoRSCLIP~\cite{zhang2024rs5m} models their similarity distributions as Gaussian functions:

\begin{align}
f^+_k(s) &= \frac{1}{\sigma^+_k \sqrt{2\pi}} \exp\left(-\frac{(s - \mu^+_k)^2}{2(\sigma^+_k)^2}\right),
\end{align}
\begin{align}
f^-_k(s) &= \frac{1}{\sigma^-_k \sqrt{2\pi}} \exp\left(-\frac{(s - \mu^-_k)^2}{2(\sigma^-_k)^2}\right),
\end{align}
where $\mu^+_k, \sigma^+_k$ and $\mu^-_k, \sigma^-_k$ denote the means and standard deviations of similarity scores for matched and mismatched pairs, respectively.

To automatically separate the two distributions,  GeoRSCLIP learns a decision boundary $t_k$ by minimizing the weighted overlap between the two distributions:

\begin{equation}
t_k = \mathop{\arg\min}_{t \ge 0} \left[ \alpha \int_{t}^{+\infty} f^-_k(s) \, ds + \int_{-\infty}^{t} f^+_k(s) \, ds \right],
\end{equation}

where $\alpha$ is a penalty parameter balancing false positives and false negatives. The learned boundary $t_k$ is then used to separate positive and negative fragments during attention calculation, enabling more robust alignment under noisy supervision.






\paragraph{Negative Branch.}
For each word $u_i$, we compute its maximum similarity with all image regions:

\begin{equation}
s_i = \max_{j} \left( \text{cos}(u_i, v_j) - t_k \right),
\end{equation}

and apply a negative mask to suppress matched words:

\begin{equation}
s^{\text{neg}}_i = s_i \cdot \text{Mask}_{\text{neg}}(s_i),
\end{equation}

where $\text{Mask}_{\text{neg}}(s_i) = 1$ if $s_i < 0$, and $0$ otherwise.

\paragraph{Positive Branch.}
We compute inter-modal attention weights:

\begin{equation}
w^{\text{inter}}_{ij} = \text{softmax} \left( \text{Mask}_{\text{pos}}(s_{ij} - t_k) \right),
\end{equation}
where $\text{Mask}_{\text{neg}}(s_i) = 1$ if $s_i > 0$, and $0$ otherwise. And aggregate matched image features to get the positive score:

\begin{equation}
s^{\text{pos}}_i = \text{cos}(u_i, \hat{v}_i) + \sum_j w^{\text{relev}}_{ij} s_{ij},
\end{equation}

where $\hat{v}_i = \sum_j w^{\text{inter}}_{ij} v_j$, and $w^{\text{relev}}_{ij}$ is a relevance-based attention weight.

\setlength{\tabcolsep}{3pt}  
\begin{table*}[!ht]
    \centering
    \label{main_reslut}
    \begin{tabular}{c|c|ccc|ccc|c|ccc|ccc|c}
    \toprule
    \hline
        \multirow{3}{*}{MRate} 
        &\multirow{3}{*}{Method}  
        &\multicolumn{7}{c|}{RSICD} 
        &\multicolumn{7}{c}{RSITMD} 
        \\ \cline{3-16}

        ~ & ~ 
        & \multicolumn{3}{c|}{Sentence Retrieval}  
        & \multicolumn{3}{c|}{Image Retrieval} 
        & \multirow{2}{*}{mR}
        & \multicolumn{3}{c|}{Sentence Retrieval}  
        & \multicolumn{3}{c|}{Image Retrieval} 
        & \multirow{2}{*}{mR} \\\cline{3-8} \cline{10-15}
        
        ~ & ~ & R1 & R5 & R10 & R1 & R5 & R10 & ~ & R1 & R5 & R10 & R1 & R5 & R10 & ~ \\ \hline

        \multirow{6}{*}{0} &  L2RM & 5.01 & 13.19 & 21.84 & 4.32 & 15.21 & 26.30 & 14.15 & 7.37 & 23.07 & 35.73 & 7.42 & 29.76 & 49.77 & 25.45 \\
        ~& HarMA-Vit & 1.49 & 3.83 & 7.53 & 1.15 & 3.47 & 6.80 & 3.97 & 2.89 & 4.82 & 9.67 & 1.61 & 4.61 & 9.63 & 5.57 \\
        ~& PIR & 8.61 & 23.57 & 35.40 & 5.49 & 21.84 & 38.04 & 21.99 & 11.14 & 33.14 & 45.21 & 8.60 & 33.02 & 56.43 & 30.99 \\
        ~& SWAN & 9.82 & 24.61 & 36.25 & 6.70 & 22.51 & 38.67 & 23.48 & 11.91 & 34.29 & 46.16 & 9.50 & 33.66 & 57.14 & 32.63 \\
        ~& DOVE & 10.54 & 25.31 & 38.27 & 7.93 & 23.54 & 40.13 & 23.79 & 13.51 & 35.25 & 48.12 & 10.21 & 34.52 & 58.45 & 33.29 \\
        ~& \textbf{PMPGuard} & \textbf{12.56} & \textbf{28.20} & \textbf{39.37} & \textbf{9.76} & \textbf{25.55} & \textbf{42.49} & \textbf{26.21} & \textbf{14.94} & \textbf{37.39} & \textbf{49.11} & \textbf{12.58} & \textbf{37.06} & \textbf{59.01} & \textbf{34.95} \\

        \hline
        
        \multirow{6}{*}{0.2} & L2RM & 4.03  & 12.35  & 21.04  & 3.59  & 14.49  & 25.51  & 13.50  & 6.64  & 22.35  & 34.96  & 6.86  & 28.94  & 48.94  & 24.78 \\ 
        ~ & HarMA-Vit & 0.82  & 3.29  & 6.50  & 0.49  & 2.71  & 5.65  & 3.24  & 2.21  & 3.98  & 8.63  & 0.88  & 3.76  & 8.89  & 4.73  \\ 
        ~ & PIR & 7.78 & 22.69 & 34.40 & 4.85 & 20.70 & 36.94 & 21.23 & 10.4 & 32.08 & 44.25 & 7.65 & 32.08 & 55.31 & 30.29 \\ 
        ~ & SWAN & 8.95 & 23.72 & 35.07 & 5.85 & 21.66 & 37.91 & 22.66 & 11.25 & 33.25 & 45.06 & 8.85 & 32.85 & 56.04 & 31.87 \\
        ~ & DOVE & 9.88 & 24.25 & 37.34 & 7.04 & 22.82 & 39.43 & 23.11 & 12.57 & 34.55 & 47.01 & 9.61 & 33.66 & 57.46 & 32.48 \\
        ~ & \textbf{PMPGuard} & \textbf{11.90} & \textbf{27.26} & \textbf{38.20} & \textbf{9.01} & \textbf{24.42} & \textbf{41.67} & \textbf{25.42} & \textbf{14.06} & \textbf{36.64} & \textbf{48.46} & \textbf{11.87} & \textbf{36.09} & \textbf{57.88} & \textbf{34.07} \\

        \hline
        \multirow{6}{*}{0.4} & L2RM & 2.84  & 9.06  & 17.11  & 2.73  & 11.40  & 20.55  & 10.61  & 5.09  & 17.92  & 30.09  & 4.47  & 20.13  & 35.49  & 18.86  \\    
        ~ & HarMA-Vit & 1.01  & 3.20  & 6.86  & 0.59  & 1.99  & 4.96  & 3.10  & 1.77  & 5.31  & 7.96  & 0.93  & 3.98  & 7.30  & 4.54  \\ 
        ~ & PIR & 6.22 & 17.38 & 29.37 & 4.41 & 18.50 & 32.81 & 18.12 & 9.51 & 26.33 & 40.27 & 7.08 & 29.56 & 51.59 & 27.39 \\ 
        ~ & SWAN & 7.44 & 22.31 & 33.60 & 4.41 & 20.29 & 36.50 & 21.19 & 9.82 & 31.77 & 43.69 & 7.33 & 31.36 & 54.47 & 30.26 \\
        ~ & DOVE & 8.35 & 22.95 & 35.96 & 5.48 & 21.31 & 37.86 & 21.77 & 11.06 & 33.17 & 45.60 & 8.21 & 32.18 & 56.01 & 30.91 \\
        ~ & \textbf{PMPGuard} & \textbf{10.38} & \textbf{25.77} & \textbf{36.90} & \textbf{7.44} & \textbf{22.92} & \textbf{40.18} & \textbf{24.03} & \textbf{12.61} & \textbf{35.29} & \textbf{47.25} & \textbf{10.41} & \textbf{34.63} & \textbf{56.62} & \textbf{32.58} \\

        \hline
        \multirow{6}{*}{0.6} & L2RM & 0.88  & 2.21  & 3.32  & 0.35  & 2.35  & 4.56  & 2.28  & 3.54  & 15.04  & 25.44  & 4.20  & 18.10  & 29.47  & 15.97  \\ 
        ~ & HarMA-Vit & 0.64  & 2.84  & 5.40  & 0.60  & 2.51  & 4.56  & 2.76  & 1.11  & 5.75  & 7.74  & 0.40  & 3.50  & 6.33  & 4.14  \\  
        ~ & PIR & 4.67 & 13.72 & 25.07 & 2.95 & 14.82 & 26.55 & 14.63 & 9.96 & 27.88 & 38.72 & 6.46 & 26.28 & 48.36 & 26.28  \\ 
        ~ & SWAN & 6.59 & 21.61 & 32.44 & 3.63 & 19.51 & 35.58 & 20.34 & 8.94 & 31.21 & 43.22 & 6.59 & 30.80 & 53.65 & 29.31 \\
        ~ & DOVE & 7.49 & 22.30 & 34.80 & 4.64 & 20.48 & 36.86 & 20.97 & 10.28 & 32.66 & 44.93 & 7.53 & 31.51 & 55.11 & 30.34 \\
        ~ & \textbf{PMPGuard} & \textbf{9.51} & \textbf{24.94} & \textbf{35.70} & \textbf{6.66} & \textbf{22.05} & \textbf{39.23} & \textbf{23.03} & \textbf{11.71} & \textbf{34.74} & \textbf{46.61} & \textbf{9.58} & \textbf{33.87} & \textbf{55.83} & \textbf{31.54} \\

        \hline
        \multirow{6}{*}{0.8} & L2RM & 0.22  & 1.55  & 3.32  & 0.58  & 2.35  & 4.56  & 2.49  & 3.54  & 11.28  & 17.92  & 1.86  & 7.88  & 14.56  & 9.51  \\ 
        ~ & HarMA-Vit & 0.27  & 1.92  & 3.66  & 0.40  & 1.70  & 3.17  & 1.85  & 0.88  & 3.32  & 4.87  & 0.35  & 2.04  & 4.96  & 2.74  \\ 
        ~ & PIR & 3.66 & 11.16 & 21.04 & 2.34 & 11.4 & 21.67 & 11.88 & 4.65 & 17.04 & 27.43 & 3.63 & 15.44 & 28.98 & 16.19 \\ 
        ~ & SWAN & 4.61 & 19.64 & 31.30 & 1.69 & 17.55 & 33.71 & 18.53 & 7.95 & 29.97 & 41.88 & 4.55 & 28.91 & 51.86 & 27.41 \\
        ~ & DOVE & 5.57 & 20.24 & 33.57 & 2.75 & 18.56 & 35.02 & 19.01 & 8.38 & 30.91 & 43.94 & 5.53 & 29.65 & 53.97 & 28.68 \\
        ~ & \textbf{PMPGuard} & \textbf{7.48} & \textbf{22.80} & \textbf{34.40} & \textbf{4.74} & \textbf{20.15} & \textbf{37.30} & \textbf{21.12} & \textbf{10.01} & \textbf{32.90} & \textbf{44.97} & \textbf{7.48} & \textbf{31.72} & \textbf{54.77} & \textbf{29.64} \\

        \bottomrule
        \bottomrule
    \end{tabular}
    \caption{Image-text retrieval performance under different MRate on RSICD and RSITMD.}
\end{table*}

\subsubsection{Positive-negative Aware Loss}

The final similarity score is computed as:

\begin{equation}
S(\mathcal{V}, \mathcal{U}) = \frac{1}{m} \sum_{i=1}^{m} \left( s^{\text{pos}}_i + s^{\text{neg}}_i \right).
\end{equation}

We adopt the bidirectional triplet ranking loss for training:

\begin{align}
\mathcal{L_{PA}} &= \sum_{(\mathcal{U}, \mathcal{V})} \left[ \gamma - S(\mathcal{U}, \mathcal{V}) + S(\mathcal{U}, \mathcal{V}') \right]_+ \nonumber \\
&+ \left[ \gamma - S(\mathcal{U}, \mathcal{V}) + S(\mathcal{U}', \mathcal{V}) \right]_+,
\end{align}

where $\gamma$ is a margin hyperparameter, and $[\cdot]_+$ denotes the hinge loss.

\subsection{The Training Objective}
The final training objective is the joint minimization of the Inter-modal Aggregation Loss and the Positive–Negative Awareness Loss:

\begin{equation}
\mathcal{L}_{\text{total}} = \mathcal{L}_{\text{IA}} + \lambda\, \mathcal{L}_{\text{PA}},
\end{equation}
where $\lambda > 0$ balances the two terms. Minimizing $\mathcal{L}_{\text{total}}$ simultaneously aligns cross-modal representations via the InfoNCE-based $\mathcal{L}_{\text{IA}}$ and suppresses the influence of noisy or mismatched pairs via the margin-based $\mathcal{L}_{\text{PA}}$, yielding robust retrieval under imperfect supervision.

\setlength{\tabcolsep}{3pt}  

\begin{table*}[!ht]
    \centering
    \label{main_GEO_reslut}
    \begin{adjustbox}{width=1\linewidth, center}
    \begin{tabular}{c|c|ccc|ccc|c|ccc|ccc|c}
    \toprule
    \hline
        \multirow{3}{*}{MRate} 
        &\multirow{3}{*}{Method}  
        &\multicolumn{7}{c|}{RSICD} 
        &\multicolumn{7}{c}{RSITMD} 
        \\ \cline{3-16}

        ~ & ~ 
        & \multicolumn{3}{c|}{Sentence Retrieval}  
        & \multicolumn{3}{c|}{Image Retrieval} 
        & \multirow{2}{*}{mR}
        & \multicolumn{3}{c|}{Sentence Retrieval}  
        & \multicolumn{3}{c|}{Image Retrieval} 
        & \multirow{2}{*}{mR} \\\cline{3-8} \cline{10-15}
        
        ~ & ~ & R1 & R5 & R10 & R1 & R5 & R10 & ~ & R1 & R5 & R10 & R1 & R5 & R10 & ~ \\ \hline
   \multirow{3}{*}{0} & L2RM-Geo & 18.84 & 42.30 & 53.65 & 14.86 & 38.60 & 55.55 & 37.42 & 27.98 & 52.80 & 62.83 & 22.95 & 55.20 & 72.34 & 48.71 \\
         ~ & HarMA-Geo & 19.51 & 43.72 & 55.68 & 15.41 & 40.06 & 57.73 & 38.90 & 29.04 & 54.56 & 65.03 & 23.82 & 56.95 & 74.61 & 50.39  \\ 
         ~ & \textbf{PMPGuard-Geo} & \textbf{20.21} & \textbf{45.05} & \textbf{57.28} & \textbf{15.93} & \textbf{41.30} & \textbf{59.39} & \textbf{40.23} & \textbf{30.12} & \textbf{56.21} & \textbf{67.00} & \textbf{24.55} & \textbf{58.75} & \textbf{76.84} & \textbf{52.01} \\ \hline
\multirow{3}{*}{0.2} & L2RM-Geo & 17.75 & 40.65 & 52.55 & 13.68 & 37.50 & 53.82 & 36.10 & 27.00 & 51.05 & 61.38 & 21.85 & 53.40 & 70.40 & 47.25 \\
         ~ & HarMA-Geo & 18.39  & 42.36  & 54.53  & 14.29  & 38.96  & 56.25  & 37.46  & 28.10  & 53.10  & 63.72  & 22.70  & 55.62  & 73.58  & 49.47  \\
        ~ & \textbf{PMPGuard-Geo} & \textbf{19.02} & \textbf{43.85} & \textbf{56.30} & \textbf{14.85} & \textbf{40.32} & \textbf{58.11} & \textbf{38.75} & \textbf{29.15} & \textbf{54.86} & \textbf{66.10} & \textbf{23.50} & \textbf{57.48} & \textbf{76.00} & \textbf{51.10} \\ \hline
\multirow{3}{*}{0.4} &  L2RM-Geo & 17.44 & 38.76 & 51.80 & 12.92 & 36.85 & 52.28 & 34.95 & 26.75 & 50.10 & 61.92 & 21.78 & 53.25 & 70.50 & 47.30 \\
        ~ & HarMA-Geo & 18.12  & 40.35  & 53.98  & 13.50  & 38.19  & 54.47  & 36.43  & 27.88  & 52.21  & 64.38  & 22.74  & 55.53  & 73.72  & 49.41  \\ 
        ~ & \textbf{PMPGuard-Geo} & \textbf{18.75} & \textbf{41.96} & \textbf{56.18} & \textbf{14.02} & \textbf{39.50} & \textbf{56.86} & \textbf{37.70} & \textbf{28.95} & \textbf{54.10} & \textbf{66.90} & \textbf{23.45} & \textbf{57.30} & \textbf{76.05} & \textbf{51.12} \\ \hline
\multirow{3}{*}{0.6} & L2RM-Geo & 16.45 & 38.45 & 51.47 & 11.75 & 33.90 & 50.90 & 33.72 & 25.40 & 48.70 & 61.80 & 19.85 & 51.20 & 67.25 & 45.65 \\
        ~ & HarMA-Geo & 17.11  & 40.07  & 53.71  & 12.28  & 35.24  & 53.14  & 35.26  & 26.55  & 50.88  & 64.16  & 20.75  & 53.50  & 70.31  & 47.69  \\ 
        ~ & \textbf{PMPGuard-Geo} & \textbf{17.68} & \textbf{41.45} & \textbf{55.60} & \textbf{12.78} & \textbf{36.55} & \textbf{55.35} & \textbf{36.50} & \textbf{27.58} & \textbf{52.80} & \textbf{66.50} & \textbf{21.50} & \textbf{55.25} & \textbf{73.00} & \textbf{49.35} \\ \hline
\multirow{3}{*}{0.8} & L2RM-Geo & 17.54 & 36.33 & 48.33 & 11.56 & 34.27 & 50.50 & 33.06 & 25.01 & 47.08 & 60.33 & 19.50 & 49.65 & 66.38 & 44.66 \\ 
         ~ & HarMA-Geo & 18.30  & 38.24  & 50.87  & 12.17  & 36.07  & 53.16  & 34.80  & 26.33  & 49.56  & 63.50  & 20.53  & 52.26  & 69.87  & 47.01  \\
        ~ & \textbf{PMPGuard-Geo} & \textbf{19.07} & \textbf{40.15} & \textbf{53.41} & \textbf{12.78} & \textbf{37.87} & \textbf{55.82} & \textbf{36.54} & \textbf{27.65} & \textbf{52.04} & \textbf{66.68} & \textbf{21.56} & \textbf{54.87} & \textbf{73.36} & \textbf{49.36} \\

        \bottomrule
        \bottomrule
    \end{tabular}
    
    \end{adjustbox}
    \caption{The image-text retrieval performance of the large model GeoRSCLIP as the backbone network under different mismatch rates (MRate) on RSICD and RSITMD.}
\end{table*}

\setlength{\tabcolsep}{1pt}  
\begin{table}[!ht]
    \centering
    \label{large_reslut}
    \begin{adjustbox}{width=1\linewidth, center}
    \begin{tabular}{c|ccc|ccc|c}
    \toprule
    \hline
        \multirow{3}{*}{Method}  
        &\multicolumn{7}{c}{RS5M} 
        \\ \cline{2-8}

        ~ 
        & \multicolumn{3}{c|}{Sentence Retrieval}  
        & \multicolumn{3}{c|}{Image Retrieval} 
        & \multirow{2}{*}{mR}\\
        
         ~ & R1 & R5 & R10 & R1 & R5 & R10 & ~  \\ \hline
         L2RM-Geo & 13.58 & 34.42 & 45.21 & 9.84 & 31.26 & 46.90 & 30.42\\
          HarMA-Geo & 14.84 & 37.13 & 48.39 & 10.96 & 33.74 & 50.12 & 32.78\\ 
          \textbf{PMPGuard-Geo} & \textbf{16.12} & \textbf{39.80} & \textbf{51.60} & \textbf{12.10} & \textbf{36.05} & \textbf{53.43} & \textbf{35.21}\\ 
        \bottomrule
        \bottomrule
    \end{tabular}
    \end{adjustbox}
    \caption{Image-text retrieval performance on RS5M.}
\end{table}

\setlength{\tabcolsep}{3pt}  
\begin{table*}[!ht]
    \centering
    \label{Ablation_reslut}
    \begin{tabular}{c|c|ccc|ccc|c|ccc|ccc|c}
    \toprule
    \hline
        \multirow{3}{*}{MRate} 
        &\multirow{3}{*}{Method}  
        &\multicolumn{7}{c|}{RSICD} 
        &\multicolumn{7}{c}{RSITMD} 
        \\ \cline{3-16}

        ~ & ~ 
        & \multicolumn{3}{c|}{Sentence Retrieval}  
        & \multicolumn{3}{c|}{Image Retrieval} 
        & \multirow{2}{*}{mR}
        & \multicolumn{3}{c|}{Sentence Retrieval}  
        & \multicolumn{3}{c|}{Image Retrieval} 
        & \multirow{2}{*}{mR} \\\cline{3-8} \cline{10-15}
        
       ~ & ~ & R1 & R5 & R10 & R1 & R5 & R10 & ~ & R1 & R5 & R10 & R1 & R5 & R10 & ~ \\ \hline
\multirow{3}{*}{0} & w/o CGA & 18.32 & 41.65 & 52.70 & 14.22 & 37.45 & 54.80 & 36.52 & 27.15 & 51.90 & 61.30 & 22.10 & 54.30 & 71.40 & 47.53 \\
~ & w/o PNAA & 19.35 & 43.20 & 55.12 & 15.12 & 39.65 & 56.90 & 38.56 & 28.45 & 53.95 & 64.20 & 23.35 & 56.40 & 73.95 & 49.38  \\ 
~ & \textbf{full} & \textbf{20.21} & \textbf{45.05} & \textbf{57.28} & \textbf{15.93} & \textbf{41.30} & \textbf{59.39} & \textbf{40.23} & \textbf{30.12} & \textbf{56.21} & \textbf{67.00} & \textbf{24.55} & \textbf{58.75} & \textbf{76.84} & \textbf{52.01} \\ \hline

\multirow{3}{*}{0.2} & w/o CGA & 17.25 & 39.45 & 50.83 & 13.10 & 36.40 & 52.70 & 34.96 & 26.10 & 49.90 & 60.20 & 20.70 & 52.30 & 69.20 & 46.57 \\
~ & w/o PNAA & 18.15 & 41.30 & 53.20 & 13.85 & 38.05 & 54.85 & 36.90 & 27.30 & 52.40 & 62.90 & 22.05 & 54.90 & 72.60 & 48.69  \\
~ & \textbf{full} & \textbf{19.02} & \textbf{43.85} & \textbf{56.30} & \textbf{14.85} & \textbf{40.32} & \textbf{58.11} & \textbf{38.75} & \textbf{29.15} & \textbf{54.86} & \textbf{66.10} & \textbf{23.50} & \textbf{57.48} & \textbf{76.00} & \textbf{51.10} \\ \hline

\multirow{3}{*}{0.4} & w/o CGA & 16.85 & 37.80 & 49.65 & 12.50 & 35.50 & 51.40 & 33.45 & 25.60 & 48.80 & 59.80 & 20.35 & 50.80 & 67.90 & 45.54 \\
~ & w/o PNAA & 17.76 & 39.90 & 52.30 & 13.20 & 37.10 & 53.20 & 35.91 & 26.65 & 51.30 & 63.30 & 21.45 & 53.30 & 71.45 & 48.24  \\ 
~ & \textbf{full} & \textbf{18.75} & \textbf{41.96} & \textbf{56.18} & \textbf{14.02} & \textbf{39.50} & \textbf{56.86} & \textbf{37.70} & \textbf{28.95} & \textbf{54.10} & \textbf{66.90} & \textbf{23.45} & \textbf{57.30} & \textbf{76.05} & \textbf{51.12} \\ \hline

\multirow{3}{*}{0.6} & w/o CGA & 15.90 & 36.40 & 48.75 & 11.30 & 32.80 & 49.85 & 32.50 & 24.15 & 47.10 & 59.60 & 18.80 & 49.85 & 65.20 & 44.12 \\
~ & w/o PNAA & 16.68 & 38.20 & 51.10 & 12.05 & 34.40 & 51.70 & 34.69 & 25.65 & 49.85 & 62.30 & 20.00 & 52.40 & 68.95 & 46.52  \\ 
~ & \textbf{full} & \textbf{17.68} & \textbf{41.45} & \textbf{55.60} & \textbf{12.78} & \textbf{36.55} & \textbf{55.35} & \textbf{36.50} & \textbf{27.58} & \textbf{52.80} & \textbf{66.50} & \textbf{21.50} & \textbf{55.25} & \textbf{73.00} & \textbf{49.35} \\ \hline

\multirow{3}{*}{0.8} & w/o CGA & 15.40 & 34.50 & 46.00 & 10.75 & 31.40 & 48.10 & 31.36 & 23.45 & 45.85 & 58.00 & 18.10 & 48.05 & 63.30 & 42.79 \\ 
~ & w/o PNAA & 16.88 & 36.80 & 49.45 & 11.80 & 33.85 & 50.65 & 33.57 & 24.95 & 48.60 & 61.30 & 19.30 & 51.20 & 66.75 & 45.02 \\
~ & \textbf{full} & \textbf{19.07} & \textbf{40.15} & \textbf{53.41} & \textbf{12.78} & \textbf{37.87} & \textbf{55.82} & \textbf{36.54} & \textbf{27.65} & \textbf{52.04} & \textbf{66.68} & \textbf{21.56} & \textbf{54.87} & \textbf{73.36} & \textbf{49.36} \\
        
        \bottomrule
        \bottomrule
    \end{tabular}
    \caption{Ablation experiments of the PMPGuard model under different mismatch rates (MRate) on RSICD and RSITMD.}
\end{table*}


        

\section{Experiments}
\subsection{Datasets}
\subsubsection{RSICD} RSICD~\cite{DBLP:journals/tgrs/LuWZL18} is a dedicated remote-sensing image-captioning benchmark that comprises 10,921 geo-diverse images—drawn from Google Earth and Street View—each paired with five concise English captions. Spanning urban, rural, agricultural, mountainous, and aquatic scenes, the dataset has been uniformly resized to 224×224 pixels to facilitate consistent processing.

\subsubsection{RSITMD} RSITMD~\cite{DBLP:journals/tgrs/YuanZFLDWS22} is a fine-grained benchmark for cross-modal remote-sensing image retrieval. It comprises 4,743 images paired with 23,715 diverse, low-redundancy captions and 1–5 precise keywords per image, yielding richer object-level descriptions. Designed to minimize intra-class similarity and amplify inter-class diversity, RSITMD delivers superior retrieval accuracy compared with existing datasets.
\subsubsection{RS5M} RS5M~\cite{zhang2024rs5m} is a five-million pair remote sensing vision language corpus whose captions were automatically scraped from open platforms. The sheer scale comes at the cost of pervasive pseudo-matched pairs (PMPs): brief, template-like, or outright irrelevant text routinely accompanies images of agriculture, infrastructure, or land-use, yielding noisy, weak, and geographically skewed alignments. A quality-scored subset reveals that up to 30\% of pairs exhibit significant mismatch, making RS5M an ideal testbed for studying retrieval robustness under realistic, imperfect supervision.
\subsection{Implementation Details}
All experiments in this study were conducted using two NVIDIA RTX A6000 GPUs. For consistency across different datasets with varying image sizes, we uniformly resized them to 224 × 224 pixels and input them into the network. Data augmentation techniques, including rotation and flip, were applied to enhance model robustness. The word vector's representation dimension was 768, while the embedding space for images and text was 512. The parameter $\tau$ was initialized to 0.07 and learned during contrast loss calculation, and a 0.5 margin was enforced for the triplet loss calculation. The network was trained for 20 epochs using the AdamW optimizer, which combined both contrast loss and triplet loss, with a batch size of 128 per GPU. To ensure the robustness of the experimental results, each set of experiments was conducted five times, and the average result was recorded.
\subsection{Baselines}
We compare PMPGuard with five state-of-the-art cross-modal retrieval methods, including five general methods: PIR~\cite{DBLP:conf/mm/PanMB23}, SWAN~\cite{pan2023reducing}, HarMA~\cite{DBLP:journals/corr/abs-2404-18253}, DOVE~\cite{ma2024direction}, Furthermore, L2RM~\cite{DBLP:conf/cvpr/HanZDL024}. L2RM and HarMa using GeoRSCLIP~\cite{zhang2024rs5m} as the backbone are treated as large-parameter models.
\subsection{Main Results}
\subsubsection{Construction of Pseudo-matched  Training Data}



To evaluate robustness under imperfect supervision, we simulate realistic noise via pseudo-matched training sets with controlled mismatches. For a clean dataset $\mathcal{D} = {(V_i, T_i)}_{i=1}^N$, we generate $\mathcal{D}_x$ by replacing $T_i$ in $x\%$ of pairs with a random $T_j$ ($j \neq i$). To ensure a true semantic mismatch, we use the pretrained \textbf{GeoRSCLIP} model to compute similarity scores and retain only replacements with scores below a threshold $\tau$. This filtering prevents trivial or semantically similar substitutions, increasing the realism of the noise. We construct multiple variants with different mismatch rates ($x = 0\%, 20\%, \dots, 80\%$) to benchmark retrieval models under varying noise levels.
\subsubsection{Results on Synthesized PMPs}

Table 1 presents the experimental results on RSICD and RSITMD, where our model adopts Swin-Transformer~\cite{DBLP:conf/iccv/LiuL00W0LG21} as the image backbone and BERT~\cite{DBLP:conf/naacl/DevlinCLT19} as the text backbone. From the results, we observe that PMPGuard achieves the best performance across all metrics compared to other state-of-the-art methods, demonstrating its strong capability for robust cross-modal retrieval. Moreover, when the mismatch ratio is higher, such as 0.6 and 0.8, the improvements brought about by PMPGuard become even more pronounced, validating the effectiveness of mismatched pair mining to enhance retrieval robustness.

\subsubsection{Evaluation Results Based on the Large Pretrained Backbone Model}
Table 2 presents the performance of different methods under varying mismatch rates (MRate) on the RSICD and RSITMD datasets. In general, our proposed method, \textbf{PMPGuard-Geo}, consistently outperforms baselines (L2RM-Geo and HarMA-Geo) in all settings and evaluation metrics. In RSICD, PMPGuard-Geo achieves the highest mR at each mismatch level, with a notable improvement from 37.42 (L2RM-Geo) and 38.90 (HarMA-Geo) to 40.23 at MRate = 0. Even as the mismatch rate increases to 0.8, PMPGuard-Geo maintains strong performance (mR = 36.54), demonstrating superior robustness to pseudo-matched pairs. On RSITMD, PMPGuard-Geo exhibits a similar advantage, consistently achieving the best R@1, R@5, R@10, and mR values. For example, at MRate = 0.8, PMPGuard-Geo obtains an mR of 49.36, significantly surpassing L2RM-Geo (44.66) and HarMA-Geo (47.01). These results validate that PMPGuard-Geo not only captures fine-grained cross-modal semantics more effectively but is also more resilient to noisy supervision, making it well-suited for real remote sensing image-text retrieval tasks.
\subsubsection{Results on Real PMPs}
Our key findings highlight the effectiveness of PMPGuard-Geo in addressing pseudo-matched pairs (PMPs) in remote sensing image-text retrieval. As shown in Table 3, PMPGuard-Geo achieves best-in-class retrieval performance, with substantial gains in R1 scores for both sentence and image retrieval, demonstrating strong precision and the ability to suppress noise via Cross-Gated Attention (CGA). Consistent improvements in R@10 further validate Positive-Negative Awareness Attention (PNAA), which effectively distinguishes latent semantic cues from noise. In particular, PMPGuard-Geo achieves a 4.79\% improvement in mean recall (mR), confirming its robustness in the retrieval directions and thresholds. Furthermore, the performance gap in the RS5M data set, compared to synthetic noise baselines, demonstrates PMPGuard's unique ability to handle real-world PMP challenges, including natural alignment inconsistencies, geographic variation, and large-scale noisy supervision. These results confirm that explicitly modeling positive-negative relationships and adaptively gating cross-modal signals transform PMPs from learning obstacles into valuable supervision.
\subsection{Ablation Experiment}
As shown in Table 4, we performed ablation studies with varying MRate in the RSICD and RSITMD datasets to evaluate the effectiveness of each component in the proposed PMPGuard-Geo framework. We examine the performance of variants without the CGA or PNAA modules (denoted as “w/o CGA” and “w/o PNAA”) and compare them with the full model. Removing either CGA or PNAA results in a consistent performance drop across all metrics and mismatch levels. In particular, excluding PNAA causes a sharper decline at higher mismatch rates: for example, mean recall (mR) on RSITMD drops from 49.36 to 45.02 at MRate = 0.8, demonstrating PNAA’s effectiveness in handling noisy supervision. Omitting \textbf{CGA} broadly reduces recall, particularly in sentence retrieval, indicating its role in enhancing cross-modal alignment. Overall, \textbf{CGA} and \textbf{PNAA} play complementary roles: CGA strengthens semantic correspondence, while PNAA improves robustness to noisy or weakly aligned pairs, jointly enabling PMPGuard-Geo to maintain 

\subsection{Visualization Experiment}
Fig. 3 visualizes PMPGuard's ability to correct mismatched remote sensing (RS) image-text pairs. Three initial incorrect caption-image match cases are presented, where PMPGuard successfully rematches each image with more appropriate descriptions—green highlights correct matches, and red denotes original mismatches. This demonstrates PMPGuard’s effectiveness in enhancing RS image-text semantic alignment and cross-modal retrieval accuracy, a key capability for real-world applications requiring precise image understanding.
\begin{figure}[t]
	\centering
	\includegraphics[width=1\linewidth]{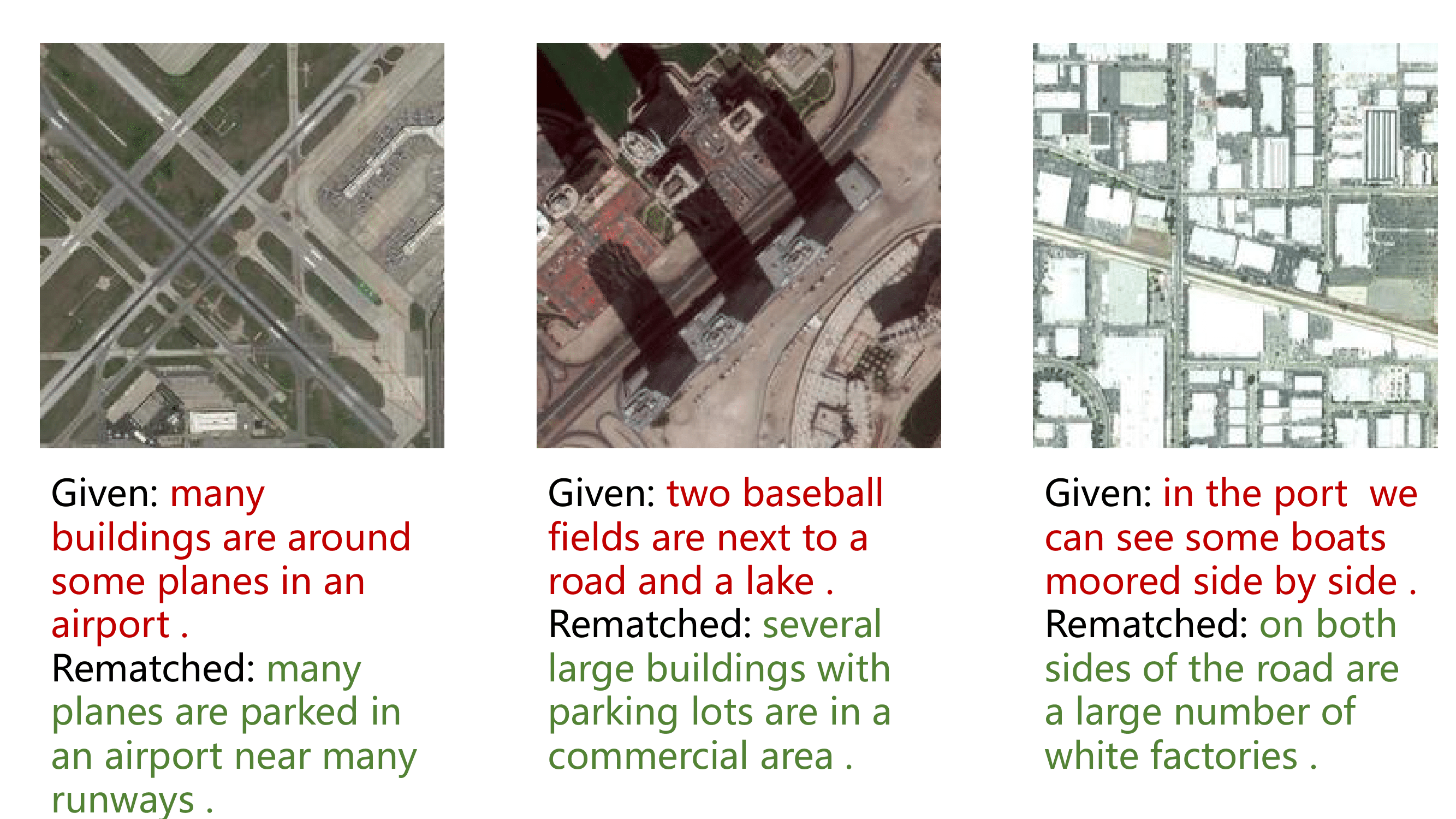}
	\caption{PMPGuard rematches mismatched RS image-text pairs; green and red words denote correct and incorrect matches, respectively.}
	\label{fig3}	
\end{figure}

\section{Conclusion}
We present \textbf{PMPGuard}, a robust remote-sensing (RS) image–text retrieval framework addressing pseudo-matched pairs (PMPs) via two core innovations: \textbf{Cross-Gated Attention (CGA)} (dynamic cross-modal feature modulation) and \textbf{Positive–Negative Awareness Attention (PNAA)} (discriminative alignment learning). Extensive experiments on RSICD, RSITMD, and RS5M demonstrate its superior performance across 0–0.8 mismatch rates, achieving SOTA results with robustness to noisy correspondences, providing a principled solution for real-world imperfect alignment retrieval scenarios and laying a foundation for robust cross-modal RS learning.

\section{Acknowledgments} This work is partially supported by the Zhejiang Provincial Natural Science Foundation of China under Grants No. LY24F020020 and No. LRG25F020002 and the Natural Science Foundation of China under Grants No. U20A20196 and No. 62302453. 
\nocite{*}
\bibliography{aaai2026}
\end{document}